# ANALYSE DE LA RIGIDITÉ DES MACHINES OUTILS 3 AXES D'ARCHITECTURE PARALLÈLE HYPERSTATIQUE

Anatol Pashkevich, Damien Chablat, Philippe Wenger
Institut de Recherche en Communications et Cybernétique de Nantes
1 rue de la Noë, 44321 Nantes
{Anatol.Pashkevich, Damien.Chablat, Philippe.Wenger}@irccyn.ec-nantes.fr

**Résumé :** Cet article présente une nouvelle méthode pour modéliser la rigidité des machines outils à structure parallèle basée sur une modélisation de la machine par des corps rigides et des flexibilités localisées. Nous ajoutons des articulations virtuelles au modèle rigide, qui décrivent les déformations élastiques des composants du mécanisme (membrures, articulations et actionneurs). Cette approche provient des travaux de Clément Gosselin, qui a évalué la rigidité des mécanismes parallèles, en tenant compte uniquement des actionneurs et qui les a modélisés comme des ressorts linéaires unidimensionnels (les membrures étaient supposées rigides, et les articulations passives parfaites). Contrairement aux études précédentes, notre méthode peut s'appliquer aux mécanismes hyperstatiques et l'évaluation des flexibilités localisées est réalisée en utilisant une modélisation par éléments finis.

**Mots clés :** Machines outils, rigidité, flexibilités localisées

**Abstract:** The paper presents a new stiffness modelling method for overconstrained parallel manipulators, which is applied to 3-d.o.f. translational mechanisms. It is based on a multidimensional lumped-parameter model that replaces the link flexibility by localized 6-d.o.f. virtual springs. In contrast to other works, the method includes a FEA-based link stiffness evaluation and employs a new solution strategy of the kinetostatic equations, which allows computing the stiffness matrix for the overconstrained architectures and for the singular manipulator postures. The advantages of the developed technique are confirmed by application examples, which deal with comparative stiffness analysis of two translational parallel manipulators.

## 1 Introduction

Comparativement aux manipulateurs sériels, les manipulateurs parallèles (MP) sont réputés pour avoir un meilleur rapport rigidité/poids et une meilleure précision. Cette caractéristique les rend attrayants pour la conception de nouvelles machines-outils pour l'usinage à grande vitesse [1, 2, 3]. Quand un manipulateur parallèle est utilisé comme machine outil, la rigidité devient une contrainte très importante lors de sa conception [4, 5, 6, 7]. Cet article présente une méthode générale pour évaluer la rigidité des manipulateurs à trois degrés de liberté hyperstatique produisant des mouvements de translation pure. Généralement, l'analyse de la rigidité est basée sur une modélisation cinétostatique [8], qui propose une carte de la rigidité en prenant en compte la rigidité des articulations motorisées. Cependant, cette méthode n'est pas appropriée pour les MP, dont les jambes sont soumises à la flexion [9].

Trois modèles existent qui permettent de prendre en compte la complaisance d'une MP afin d'analyser son comportement flexible : les méthodes par éléments finis (MEF) [10], modélisation avec membrures rigides (MMR) [11], et modélisation rigide avec flexibilités localisées (MRFL) [8 ]. Cette dernière méthode est basée sur l'expansion du modèle traditionnel rigide en ajoutant les articulations virtuelles, qui décrivent les déformations élastiques des composants du mécanisme (membrures, articulations et actionneurs). Cette approche provient des travaux de Gosselin [8], qui a évalué la rigidité des mécanismes parallèles, en tenant compte uniquement des actionneurs et qui les a modélisés comme des ressorts linéaires unidimensionnels (les membrures étaient supposées rigides, et les articulations passives parfaites). Ces hypothèses ont permis de réduire l'analyse de la rigidité d'un mécanisme à l'analyse de sa matrice jacobienne. Une évolution de ce modèle, tenant compte des flexibilités dans les membrures a été présentée dans [10] en utilisant des corps rigides et des flexibilités par des ressorts en torsion ou en flexion. Généralement, cette modélisation fournit une précision suffisante et les temps de calcul sont courts. On peut donc l'utiliser en phase de conception préliminaire, en particulier pour les analyses paramétriques. Cependant, nous verrons que des hypothèses de simplification qui ne prennent pas en compte le couplage entre la translation et de rotation peuvent aboutir à d'importantes erreurs. De plus, il existe également d'autres restrictions, qui limitent ses applications à l'absence de mécanismes hyperstatiques.

Dans le paragraphe suivant, nous présentons une méthode générale pour déduire la modèle cinématique et le modèle de raideur. Le paragraphe 3 décrit les matrices de complaisance de chaque élément et leur assemblage pour déterminer la rigidité globale du mécanisme. Finalement, dans le paragraphe 4, nous appliquons notre méthode sur deux exemples.

## 2 Méthodologie

### 2.1 Architecture des manipulateurs étudiés

Soit un manipulateur parallèle général à 3 degrés de liberté en translation, possédant une plate-forme mobile connectée à une base fixe par trois chaînes cinématiques identiques (Fig. 1).

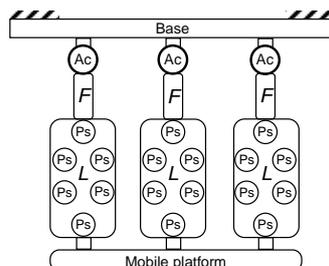

*Fig. 1. Schéma d'ordre général d'un manipulateur parallèle à 3-ddl de translation*
*(Ac – articulation motorisée, Ps – articulation passive, F - pied, L – jambe)*

Chaque jambe comprend une articulation motorisée "Ac" (prismatique ou rotoïde), suivi d'un pied et d'une jambe possédant un certain nombre d'articulation passive "P". Des conditions géométriques existent sur les articulations passivent pour permettre d'éliminer les mouvements de rotation de la plate-forme et d'assurer ainsi la stabilité de ses mouvements en translation. Nous présentons trois exemples de ces manipulateurs :

- le robot 3-PUU (Figure 2a), où chaque jambe est composée d'une tige terminée par deux cardans (avec des conditions de parallélismes entre les axes des cardans), et des articulations prismatiques motorisées [13] ;
- le robot Delta (Figure 2b), qui est une architecture de type 3-RRPaR avec un parallélogramme dans chaque jambe et des articulations rotoïdes [14] ;
- le robot parallèle Orthoglide (Fig 2c), qui est une architecture de type 3-PRPaR avec un parallélogramme dans chaque jambe et des articulations prismatique motorisées [10].

Les lettres R, P, U et Pa représentent les articulations rotoïdes, prismatiques, les cardans et les parallélogrammes, respectivement. Les exemples (b) et (c) illustrent des mécanismes hyperstatiques, où certaines contraintes cinématiques sont redondantes, mais n'affectent pas le nombre de degrés de liberté. Cependant, la plupart de ces méthodes d'analyse de la rigidité des manipulateurs parallèles ne s'applique qu'à des manipulateurs isostatique [8].

*(a) Robot 3-PUU [13]*      *(b) Robot Delta[14]*      *(c) Robot Orthoglide [10]*

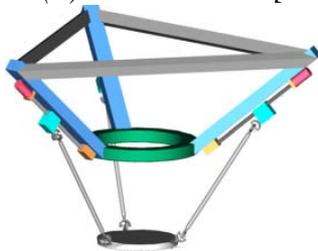 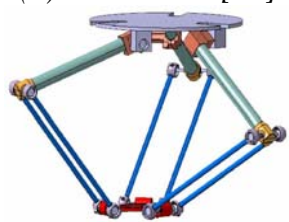 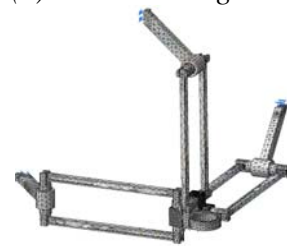

*Fig. 2. Exemples de manipulateur parallèle à 3 ddl*

## 2.2 Hypothèses de base

Pour évaluer la rigidité d'un manipulateur, nous appliquons une petite modification à la méthode des travaux virtuelle (MTV), qui est basée sur la méthode des flexibilités localisées [8, 10]. Selon cette approche, le modèle rigide de départ est modifié en ajoutant des articulations virtuelles, qui décrivent les déformations élastiques des corps rigides. De plus, des ressorts virtuels sont inclus dans les articulations motorisées pour tenir compte de la rigidité de la boucle de contrôle. Pour surmonter les difficultés de modélisation des parallélogrammes, nous allons dans un premier temps les remplacer (voir Fig. 2) par des corps rigides dont la rigidité dépend de la configuration. Cette modification transforme l'architecture générale en un mécanisme 3-xUU et nous permet ainsi d'étudier plusieurs manipulateurs de cette famille. Avec cette hypothèse, chaque chaîne cinématique du manipulateur peut être décrite par une structure sérielle (Fig. 3), qui comprend successivement:

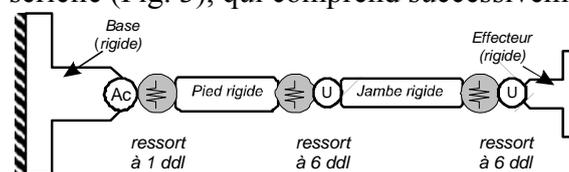

*Fig. 3. Modèle flexible d'une seule chaîne cinématique*

- un corps rigide entre la base fixe du manipulateur et la $i^{\text{ème}}$ articulation motorisée décrite par une matrice de transformation homogène constante $\mathbf{T}_{base}^{i}$ ;
- une articulation motorisée avec un degré de liberté avec flexibilité localisée, qui est décrite par une matrice homogène $\mathbf{V}_a(q_0^i + \theta_0^i)$ où $q_0^i$ sont les coordonnées des actionneurs et $\theta_0^i$ les coordonnées de l'articulation virtuelle ;
- un corps rigide, le pied, reliant les actionneurs de la jambe et la jambe, qui est décrite par une matrice de transformation homogène constante $\mathbf{T}_{foot}$ ;

- une articulation virtuelle à 6 degrés de liberté définissant trois mouvement de translation et de rotations, qui sont décrits en fonction d'une matrice homogène $\mathbf{V}_s(\theta_1^i,...\theta_6^i)$, où $\{\theta_1^i,\theta_2^i,\theta_3^i\}$ et $\{\theta_4^i,\theta_5^i,\theta_6^i\}$ correspondent aux translations et rotations élémentaires, respectivement ;
- une articulation passive de type cardan à 2 degrés de liberté placée à l'origine de la jambe permettant deux rotations d'angles $\{q_1^i,q_2^i\}$, qui est décrite par la matrice de transformation homogène $\mathbf{V}_{u1}(q_1^i,q_2^i)$ ;
- un corps rigide, la jambe, reliant le pied de la plate-forme mobile, qui est décrite par la matrice de transformation constante homogène $\mathbf{T}_{leg}$ ;
- une articulation virtuelle à 6 degrés de liberté définissant trois translations et trois rotations de la flexibilité localisée, qui sont décrits en fonction de la matrice homogène $\mathbf{V}_s(\theta_7^i,...\theta_{12}^i)$, où $\{\theta_7^i,\theta_8^i,\theta_9^i\}$ et $\{\theta_{10}^i,\theta_{12}^i,\theta_{12}^i\}$ correspondent aux translations et aux rotations élémentaires, respectivement ;
- une articulation passive de type cardan à 2 degrés de liberté placée à l'extrémité de la jambe permettant deux rotations d'angles $\{q_3^i,q_4^i\}$, qui est décrite par la matrice de transformation homogène $\mathbf{V}_{u2}(q_3^i,q_4^i)$ ;
- d'un lien rigide entre la jambe et l'effecteur (partie de la plate-forme mobile) décrite par la matrice de transformation homogène constante $\mathbf{T}_{tool}^i$ ;

La position de l'effecteur est soumise aux variations de toutes les coordonnées d'une seule chaîne cinématique qui peut être écrite de la manière suivante :

$$\mathbf{T}_i = \mathbf{T}_{base}^i \mathbf{V}_a(q_0^i + \theta_0^i) \mathbf{T}_{foot} \mathbf{V}_s(\theta_1^i,...\theta_6^i) \cdot \mathbf{V}_{u1}(q_1^i,q_2^i) \mathbf{T}_{leg} \mathbf{V}_s(\theta_7^i,...\theta_{12}^i) \mathbf{V}_{u2}(q_3^i,q_4^i) \mathbf{T}_{tool}^i \qquad (1)$$

où la matrice $\mathbf{V}_a(.)$ est soit une fonction élémentaire de rotations ou de translations, les matrices $\mathbf{V}_{u1}(.)$ et $\mathbf{V}_{u2}(.)$ sont des composantes de deux rotations successives, et la matrice $\mathbf{V}_s(.)$ est composée de six transformations élémentaires. Dans le cas de corps rigides, les coordonnées des articulations virtuelles sont égales à zéro, tandis que les autres angles (actifs ou passives) sont obtenus en utilisant le modèle géométrique inverse. Des expressions particulières de tous les composants du produit (1) peuvent être facilement obtenues à l'aide des techniques standards utilisés pour définir les matrices de transformation homogènes. Il convient de noter que le modèle cinématique (1) comprend 18 variables (1 pour l'articulation active et, 4 pour les articulations passive, et 13 pour les ressorts virtuels). Toutefois, certaines flexibilités localisées sont redondantes, car elles sont compensées par des articulations passives correspondant à l'alignement des axes ou par combinaison d'articulations passives. Néanmoins, il n'est pas nécessaire de les détecter ni de les supprimer pour permettre notre analyse, parce que notre technique permet facilement et efficacement d'éliminer cette redondance.

## 2.3 Modèle cinématique

Pour évaluer les capacités du manipulateur pour répondre aux forces et aux couples extérieurs, nous écrivons les équations différentielles qui décrivent les relations entre la position de l'effecteur et les petites variations des articulations. Pour chaque $i^{ème}$ chaîne cinématique, cette équation peut être généralisée comme suit

$$\delta \mathbf{t}_i = \mathbf{J}_\theta^i \cdot \delta \mathbf{\theta}_i + \mathbf{J}_q^i \cdot \delta \mathbf{q}_i, \quad i = 1,2,3 \qquad (2)$$

où le vecteur $\delta\mathbf{t}_i = (\delta p_{xi}, \delta p_{yi}, \delta p_{zi}, \delta \varphi_{xi}, \delta \varphi_{yi}, \delta \varphi_{zi})^T$ décrit la translation $\delta\mathbf{p}_i = (\delta p_{xi}, \delta p_{yi}, \delta p_{zi})^T$ et la rotation $\delta\mathbf{\varphi}_i = (\delta \varphi_{xi}, \delta \varphi_{yi}, \delta \varphi_{zi})^T$ de l'effecteur dans le repère cartésien; le vecteur $\delta\mathbf{\theta}_i = (\delta\theta_0^i, ... \delta\theta_{12}^i)^T$ comprend toutes les coordonnées des articulations virtuelles, le vecteur $\delta\mathbf{q}_i = (\delta q_1^i,...\delta q_4^i)^T$ comprend toutes les coordonnées des articulations passives, le symbole 'δ' représente les variations par rapport au cas rigide, $\mathbf{J}_\theta^i$ et $\mathbf{J}_q^i$ sont des matrices de tailles 6×13 et 6×4 respectivement. Il convient de noter que la dérivée des coordonnées articulaires motorisées $q_0^i$ n'est pas inclus dans $\mathbf{J}_q^i$ mais elle est représentée dans la première colonne de $\mathbf{J}_\theta^i$ à travers la variable $\theta_0^i$. Les matrices $\mathbf{J}_\theta^i$, $\mathbf{J}_q^i$, qui sont les seuls paramètres de l'équation différentielle (2), peuvent être calculées à partir de l'équation (1) de manière analytique en utilisant un logiciel de calcul formel comme Maple, MathCAD ou

Mathematica. Toutefois, une simple différenciation de ces expressions donne des expressions complexes à manipuler pour des calculs supplémentaires. D'autre part, la décomposition de (1), où toutes les variables sont séparées, nous permet d'appliquer des méthodes de décomposition semi-analytique. Pour présenter cette technique, nous supposons que les articulations virtuelles $\theta_0^i$, le modèle (1) peut être réécrit par,

$$\mathbf{T}_i = \mathbf{H}_{ij}^1 \cdot \mathbf{V}_{\theta j}(\theta_j^i) \cdot \mathbf{H}_{ij}^2 \tag{3}$$

où la première et la troisième matrices sont des matrices de homogènes constantes, et le second est une matrice de transformation homogène comprenant des translations et de rotations élémentaires. Ensuite, les dérivées partielles de la matrice homogène $\mathbf{T}_i$ pour les variables $\theta_j^i$ pour $\theta_j^i = 0$ peuvent être calculées à partir d'un produit similaire où le terme $\mathbf{V}'_{\theta j}(.)$ est remplacé par son expression analytique. En particulier, pour les translations et les rotations élémentaires autour de l'axe X, ces matrices peuvent s'écrire :

$$\mathbf{V}'_{Tran_x} = \begin{bmatrix} 0 & 0 & 0 & 1 \\ 0 & 0 & 0 & 0 \\ 0 & 0 & 0 & 0 \\ 0 & 0 & 0 & 0 \end{bmatrix}; \quad \mathbf{V}'_{Rot_x} = \begin{bmatrix} 0 & 0 & 0 & 0 \\ 0 & 0 & 1 & 0 \\ 0 & -1 & 0 & 0 \\ 0 & 0 & 0 & 0 \end{bmatrix}. \tag{4}$$

En outre, à partir de la dérivation de la matrice homogène, $\mathbf{T}'_i = \mathbf{H}_{ij}^1 \cdot \mathbf{V}'_{\theta j}(\theta_j^i) \cdot \mathbf{H}_{ij}^2$ peut être présentées par

$$\mathbf{T}'_i = \begin{bmatrix} 0 & \varphi'_{iz} & -\varphi'_{iy} & p'_{ix} \\ -\varphi'_{iz} & 0 & \varphi'_{ix} & p'_{iy} \\ \varphi'_{iy} & -\varphi'_{ix} & 0 & p'_{iz} \\ 0 & 0 & 0 & 0 \end{bmatrix} \tag{5}$$

Alors la j$^{\text{ème}}$ colonne de $\mathbf{J}_\theta^i$ peut être extraite à partir de $\mathbf{T}'_i$ (en utilisant les éléments des matrices $T'_{14}$, $T'_{24}$, $T'_{34}$, $T'_{23}$, $T'_{31}$, $T'_{12}$). Les matrices jacobiennes $\mathbf{J}_q^i$ peuvent être calculée de la même manière, mais elles sont évaluées au voisinage de leurs position nominales $q^i_{j_{nom}}$ qui correspond au cas rigide (ces valeurs sont fournies par le modèle cinématique inverse).
Toutefois, la simple transformation $q_j^i = q^i_{j_{nom}} + \delta q_j^i$ et la factorisation de $\mathbf{V}_{qj}(q_j^i) = \mathbf{V}_{qj}(q^i_{j_{nom}}) \cdot \mathbf{V}_{qj}(\delta q_j^i)$ permettent l'application de l'approche ci-dessus. Il convient également de mentionner que cette technique peut être utilisée dans des calculs analytiques et permet d'éviter des équations volumineuses produites par une simple différenciation automatique.

### 2.4 Modélisation de la rigidité d'un manipulateur

Pour le modèle de rigidité, qui décrit la relation entre les efforts et le déplacement, il est nécessaire d'introduire de nouvelles équations qui définissent les réactions des articulations virtuelles aux déformations des flexibilités localisées. Conformément à notre modèle de raideur, trois types de flexibilités localisées sont inclus dans chaque chaîne cinématique :
- une flexibilité localisée à 1-ddl décrivant la complaisance de l'actionneur ;
- une flexibilité localisée à 6-ddl décrivant la complaisance du pied ;
- une flexibilité localisée à 6-ddl décrivant la complaisance de la jambe.

En supposant que les déformations des flexibilités localisées sont petites, les relations peuvent être exprimées par des équations linéaires.

$$\left[\tau_{\theta 0}^i\right] = K_{act}\left[\theta_0^i\right]; \quad \begin{bmatrix} \tau_{\theta 1}^i \\ \vdots \\ \tau_{\theta 6}^i \end{bmatrix} = \mathbf{K}_{Foot} \begin{bmatrix} \theta_1^i \\ \vdots \\ \theta_6^i \end{bmatrix}; \quad \begin{bmatrix} \tau_{\theta 7}^i \\ \vdots \\ \tau_{\theta 12}^i \end{bmatrix} = \mathbf{K}_{Leg} \begin{bmatrix} \theta_7^i \\ \vdots \\ \theta_{12}^i \end{bmatrix} \tag{6}$$

où $\tau_{\theta j}^i$ est la force généralisée de la j$^{\text{ème}}$ articulation virtuelle de la i$^{\text{ème}}$ chaîne cinématique, $K_{act}$ est la raideur de l'actionneur, et $\mathbf{K}_{Foot}$, $\mathbf{K}_{Leg}$ sont des matrices de raideur de dimensions 6×6 associées aux jambes et aux pieds respectivement. Il convient de souligner que, contrairement à d'autres études, ces matrices sont supposées être non-diagonales. Cela nous permet de prendre en compte les couplages entre les déformations angulaires et de translations, qui sont souvent ignorées [8]. Pour facilité l'analyse, les expressions (6) peuvent être regroupées dans une seule matrice sous la forme,

$$\boldsymbol{\tau}_\theta^i = \mathbf{K}_\theta \cdot \delta\boldsymbol{\theta}_i, \quad i=1,2,3 \tag{7}$$

où $\boldsymbol{\tau}_\theta^i = (\tau_{\theta 0}^i, \ldots \tau_{\theta 12}^i)^T$ est le vecteur regroupant les réactions des articulations virtuelles, et $\mathbf{K}_\theta = \mathrm{diag}(K_{act}, \mathbf{K}_{Foot}, \mathbf{K}_{Leg})$ est la matrice de rigidité des flexibilités localisées de dimension 13×13. De la même manière, on peut définir le vecteur regroupant les réactions des articulations passives $\boldsymbol{\tau}_q^i = (\tau_{q1}^i, \ldots \tau_{q4}^i)^T$, mais tous ses éléments doivent être égaux à zéro :

$$\boldsymbol{\tau}_q^i = \mathbf{0}, \quad i=1,2,3 \tag{8}$$

Pour trouver les équations correspondant au déplacement de l'effecteur $\delta\mathbf{t}_i$, nous appliquons le principe des travaux virtuels en supposant que les déplacement des articulations sont petites et que ces déplacements ($\Delta\boldsymbol{\theta}_i, \Delta\mathbf{q}_i$) sont autour de la position d'équilibre. Ensuite, le travail virtuel de la force extérieure $\mathbf{f}_i$ appliquée sur l'effecteur $\Delta\mathbf{t}_i = \mathbf{J}_\theta^i \cdot \Delta\boldsymbol{\theta}_i + \mathbf{J}_q^i \cdot \Delta\mathbf{q}_i$ est égale à la somme $(\mathbf{f}_i^T \mathbf{J}_\theta^i) \cdot \Delta\boldsymbol{\theta}_i + (\mathbf{f}_i^T \mathbf{J}_q^i) \cdot \Delta\mathbf{q}_i$. Pour les forces internes, le travail virtuel est $-\boldsymbol{\tau}_\theta^{iT} \cdot \Delta\boldsymbol{\theta}_i$ puisque les articulations passives ne produisent pas la force ni de couples résistants (le signe moins prend en compte les orientations adoptées pour la définition des flexibilités localisées). Par conséquent, pour avoir un équilibre statique, les travaux virtuels doivent être égaux à zéro pour les articulations virtuel. Les conditions d'équilibre peuvent alors s'écrire

$$\mathbf{J}_\theta^{iT} \cdot \mathbf{f}_i = \boldsymbol{\tau}_\theta^i; \quad \mathbf{J}_q^{iT} \cdot \mathbf{f}_i = \mathbf{0}. \tag{9}$$

Cela donne d'autres expressions décrivant la transformation des forces et des couples entre les articulations et l'effecteur. Ainsi, le modèle cinétostatique complet se compose de cinq matrices (2), (7)…(9) où $\mathbf{f}_i$ et $\delta\mathbf{t}_i$ sont considérés comme connus, et les autres variables sont considérées comme inconnues. De toute évidence, puisque les chaînes cinématiques sont distinctes et possèdent certains degrés de liberté, ce système ne peut pas être résolu uniquement en connaissant $\mathbf{f}_i$. Cependant, pour un déplacement donné de l'effecteur $\delta\mathbf{t}_i$, il est possible de calculer les forces externes correspondants $\mathbf{f}_i$ et de les variables internes $\delta\boldsymbol{\theta}_i$, $\boldsymbol{\tau}_\theta^i$, $\delta\mathbf{q}_i$ (c'est-à-dire les déplacements de flexibilités localisées et les déplacements des articulations passive). Comme cette matrice est non singulière, $\delta\boldsymbol{\theta}_i$ peut être exprimée à partir de $\mathbf{f}_i$ à l'aide des équations $\boldsymbol{\tau}_\theta^i = \mathbf{K}_\theta \cdot \delta\boldsymbol{\theta}_i$ et $\mathbf{J}_\theta^{iT} \cdot \mathbf{f}_i = \boldsymbol{\tau}_\theta^i$. Cela nous permet de réduire notre système à cette équation

$$(\mathbf{J}_\theta^i \mathbf{K}_\theta^{-1} \mathbf{J}_\theta^{iT}) \cdot \mathbf{f}_i + \mathbf{J}_q^i \cdot \delta\mathbf{q}_i = \delta\mathbf{t}_i; \quad \mathbf{J}_q^{iT} \cdot \mathbf{f}_i = \mathbf{0} \tag{10}$$

avec les inconnues $\mathbf{f}_i$ et $\Delta\mathbf{q}_i$. Ce système peut également être réécrit sous forme de matrice

$$\begin{bmatrix} \mathbf{S}_\theta^i & \mathbf{J}_q^i \\ \mathbf{J}_q^{iT} & \mathbf{0} \end{bmatrix} \cdot \begin{bmatrix} \mathbf{f}_i \\ \delta\mathbf{q}_i \end{bmatrix} = \begin{bmatrix} \delta\mathbf{t}_i \\ \mathbf{0} \end{bmatrix} \tag{11}$$

où $\mathbf{S}_\theta^i = \mathbf{J}_\theta^i \mathbf{K}_\theta^{-1} \mathbf{J}_\theta^{iT}$ décrit le complaisance par rapport à l'effecteur, et $\mathbf{J}_q^i$ tient compte des articulations passives sur le mouvement de l'effecteur. Par conséquent, pour une chaîne cinématique, la matrice de raideur $\mathbf{K}_i$ définissant la relation entre les efforts extérieurs et les déplacements de l'effecteur est

$$\mathbf{f}_i = \mathbf{K}_i \cdot \delta\mathbf{t}_i \tag{12}$$

Cette matrice peut être calculée en inversion la matrice de dimension 10×10 en extrayant une sous-matrice de dimension 6×6 qui correspondant à $\mathbf{S}_\theta^i$. Il est également important de mentionner que notre méthode ne nécessite que l'inversion d'une 6×6 puise que $\mathbf{K}_\theta^{-1} = \mathrm{diag}(K_{act}^{-1}, \mathbf{K}_{Foot}^{-1}, \mathbf{K}_{Leg}^{-1})$.

Dans le cas général c'est-à-dire pour tout $\mathbf{J}_\theta^i$ et $\mathbf{J}_q^i$, il n'est pas possible de prouver que l'équation (11) est inversible. De plus, si $\mathbf{J}_q^i$ est singulière, les coordonnées des articulations passives $\mathbf{q}_i$ ne sont pas uniques. D'un point de vue physique, cela signifie que si la chaîne cinématique est située dans une posture singulière, certains déplacements peuvent être générés par un déplacement infinitésimal des articulations passives. Mais pour un effort donné $\mathbf{f}_i$, la solution correspondante est unique (puisque $\mathbf{J}_\theta^i$ est non singulière s'il existe au moins une flexibilité localisée dans une chaîne cinématique). D'autre part, une configuration singulière peut être associée à une infinité de matrices de raideur pour la même localisation de l'effecteur et pour différentes $\mathbf{q}_i$ calculés par le modèle géométrique inverse. Pour résoudre ce problème nous avons programmé une méthode basée sur une décomposition en valeurs singulières. Lorsque la matrice de raideur de chaque jambe $\mathbf{K}_i$ a été évaluée, nous obtenons la matrice de raideur du manipulateur par une simple addition

$$\mathbf{K}_m = \sum_{i=1}^{3} \mathbf{K}_i \qquad (13)$$

Ceci découle du principe de superposition, parce que la force extérieure totale correspondant au déplacement de l'effecteur $\delta\mathbf{t}$ (la même pour toutes les chaînes cinématiques) peut être exprimé sous la forme $\mathbf{f} = \sum_{i=1}^{3} \mathbf{f}_i$ avec $\mathbf{f}_i = \mathbf{K}_i \cdot \delta\mathbf{t}$.

Il convient de souligner que la matrice résultant n'est pas inversible, puisque certains mouvements de l'effecteur ne produisent pas les réactions sur les flexibilités localisées (à cause de l'influence des articulations passives). Cependant, pour l'ensemble de manipulateur, la matrice de raideur est définie positive et inversible pour toutes les postures non singulières.

## 3 Définitions des éléments complaisants

Pour chaque jambe, notre modèle de rigidité comprend trois éléments, qui sont décrits par une flexibilité localisée à 1ddl et deux autres à 6-ddl qui correspondes à la complaisances de l'actionneur, du pied et de la jambe (voir Fig. 3). Nous allons maintenant décrire une méthode pour évaluer les matrices de complaisance.

### 3.1 Complaisance des actionneurs

La complaisance de l'actionneur est un scalaire qui dépend de la mécanique et de la boucle d'asservissement. Comme la plupart des actionneurs sont commandés par un PID numérique, la principale source d'erreur est liée à la transmission mécanique qui souvent en dehors de la boucle de contrôle (vis à billes, engrenages, courroies, ...) dont la complaisance est comparable à la complaisance des autres éléments du manipulateur. En raison de la complexité de la structure mécanique des servomoteurs, ce paramètre est souvent évalué à partir des expériences en statique.

### 3.2 Complaisance des membrures

Dans notre étude, la complaisance des membrures ou son inverse, la rigidité, est définie par une matrice symétrique définie positive de dimension 6×6 correspondant à une flexibilité localisée à 6-ddl avec un couplage entre les déformations de translation et de rotation. Le moyen le plus simple pour obtenir ces matrices est de les assimiler à des poutres pour lesquels les éléments non-nul de la matrice de raideur sont :

$$k_{11} = \frac{L}{EA}\,;\ k_{22} = \frac{L^3}{3EI_z}\,;\ k_{33} = \frac{L^3}{3EI_y}\,;\ k_{44} = \frac{L}{GJ}\,;\ k_{55} = \frac{L}{EI_y}\,;\ k_{66} = \frac{L}{EI_z}\,;\ k_{35} = -\frac{L^2}{2EI_y}\,;\ k_{26} = \frac{L^2}{2EI_z} \quad (14)$$

avec $L$ la longueur de la poutre, $A$ sa surface de la section transversale, $I_y$, $I_z$ et $J$ sont les inerties quadratiques et polaire de la section transversale, et $E$ et $G$ sont les modules d'Young et de Coulomb, respectivement. Cependant, pour certaines géométries complexes des membrures, une simple approximation par une poutre n'est pas suffisante. Dans ce cas, nous pouvons assimiler une membrure à une série de poutres assemblées en série par encastrement et à utiliser la matrice de rigidité résultante.

### 3.3 Complaisance des membrures évaluée par une modélisation par éléments finis

Pour des membrures possédant une forme complexes, des résultats plus précis peuvent être obtenus en utilisant une modélisation par éléments finis. Pour appliquer cette approche, nous ajoutons au modèle CAO de chaque membrure, une pièce de référence "pseudo-rigide", qui est utilisé comme référence pour l'évaluation de la complaisance et l'origine de chaque membrure est fixe. Puis, en appliquant des efforts et des couples unitaires sur la pièce de référence, il est possible d'évaluer les déplacements angulaires et linéaires, ce qui permet l'évaluation des colonnes de la matrice de rigidité. La principale difficulté ici est d'obtenir les valeurs exactes des déplacements en utilisant un bon maillage. De plus, pour accroître la précision, les déplacements doivent être évalués en utilisant des données redondantes qui décrivent le déplacement du corps de référence. C'est pour cela que nous une décomposition en valeur singulière. À partir de notre étude, nous avons constaté qu'une approximation du pied par une seule poutre donne une précision de 50%, et les quatre poutres améliore jusqu'à 30% seulement. Cependant, bien que la méthode basée sur des éléments finis soit la plus précise elle est aussi la coûteuse en temps de calcul. Toutefois, contrairement à une simple modélisation par élément finis de l'ensemble du manipulateur, qui exige un nouveau maillage des conditions aux limites à chaque fois que nous changeons de posture, notre méthode ne demande qu'un calcul d'évaluation de la raideur des membrures.

## 4 Exemples d'application

Pour démontrer l'efficacité de notre méthode, nous allons l'appliquer pour comparer la rigidité de deux mécanismes à 3ddl basés sur l'architecture de l'Orthoglide. Les modèles CAO de ces mécanismes sont représentés dans la Fig. 4.

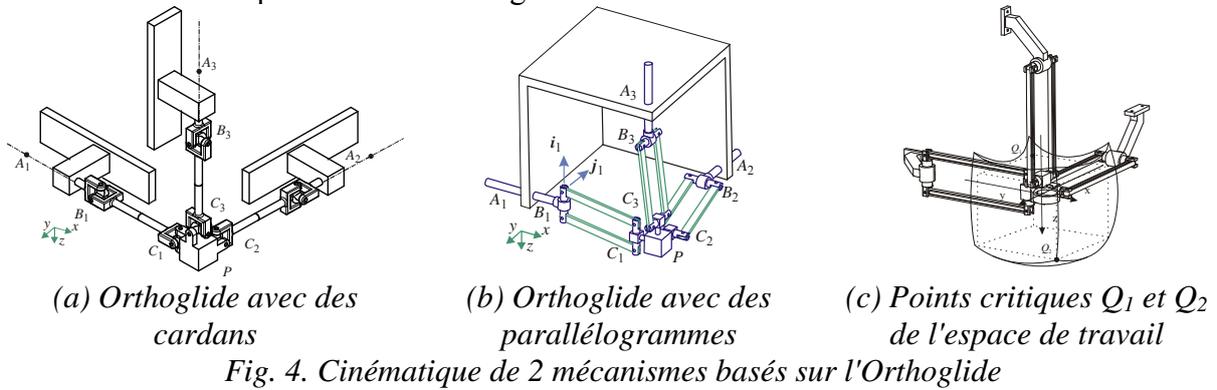

*(a) Orthoglide avec des cardans*  *(b) Orthoglide avec des parallélogrammes*  *(c) Points critiques $Q_1$ et $Q_2$ de l'espace de travail*

*Fig. 4. Cinématique de 2 mécanismes basés sur l'Orthoglide*

### 4.1 Rigidité de l'Orthoglide avec des cardans

Nous écrivons un modèle de rigidité simplifiée de l'Orthoglide où les jambes sont de type PUU c'est-à-dire avec deux cardans. Cependant, pour conserver les propriétés de l'Orthoglide, la section de la membrure entre les deux cardans est doublée pour être équivalent aux bars des parallélogrammes. En faisant l'hypothèse que l'origine de notre système de coordonnée est situé sur le repère de l'effecteur lorsque le manipulateur est en posture isotrope (quand les jambes sont mutuellement perpendiculaires et parallèles aux axes de prismatiques). Avec cette hypothèse, les modèles géométriques de différentes chaînes cinématiques peuvent être décrit par l'expression (1), où $i \in \{x, y, z\}$ et les différentes matrices sont définies par les opérateurs de translation et de rotation $\mathbf{T}_x(.), \mathbf{T}_y(.), \ldots \mathbf{R}_z(.)$ suivants :

$$\mathbf{T}_{base}^{x} = \begin{bmatrix} 1 & 0 & 0 & -L-r \\ 0 & 1 & 0 & 0 \\ 0 & 0 & 1 & 0 \\ \hline 0 & 0 & 0 & 1 \end{bmatrix} \quad \mathbf{T}_{tool}^{x} = \begin{bmatrix} 1 & 0 & 0 & r \\ 0 & 1 & 0 & 0 \\ 0 & 0 & 1 & 0 \\ \hline 0 & 0 & 0 & 1 \end{bmatrix} \quad \mathbf{T}_{base}^{y} = \begin{bmatrix} 0 & 0 & 1 & 0 \\ 1 & 0 & 0 & -L-r \\ 0 & 1 & 0 & 0 \\ \hline 0 & 0 & 0 & 1 \end{bmatrix} \quad (15)$$

$$\mathbf{T}_{tool}^{y} = \begin{bmatrix} 0 & 1 & 0 & r \\ 0 & 0 & 1 & 0 \\ 1 & 0 & 0 & 0 \\ \hline 0 & 0 & 0 & 1 \end{bmatrix} \quad \mathbf{T}_{base}^{z} = \begin{bmatrix} 0 & 1 & 0 & 0 \\ 0 & 0 & 1 & 0 \\ 1 & 0 & 0 & -L-r \\ \hline 0 & 0 & 0 & 1 \end{bmatrix} \quad \mathbf{T}_{base}^{z} = \begin{bmatrix} 0 & 0 & 1 & r \\ 1 & 0 & 0 & 0 \\ 0 & 1 & 0 & 0 \\ \hline 0 & 0 & 0 & 1 \end{bmatrix} \quad (16)$$

$$\mathbf{V}_a(q_0 + \theta_0) = \mathbf{T}_x(q_0 + \theta_0) \, ; \quad \mathbf{T}_{Foot} = \mathbf{I} \, ; \quad \mathbf{T}_{leg} = \mathbf{T}_x(L) \, ; \quad (17)$$

$$\mathbf{V}_s(\theta_1, \ldots \theta_6) = \mathbf{T}_x(\theta_1) \mathbf{T}_y(\theta_2) \mathbf{T}_z(\theta_3) \mathbf{R}_x(\theta_4) \mathbf{R}_y(\theta_5) \mathbf{R}_z(\theta_6) \quad (18)$$

$$\mathbf{V}_{u1}(q_1, q_2) = \mathbf{R}_z(q_1) \mathbf{R}_y(q_2) \, ; \quad \mathbf{V}_{u2}(q_3, q_4) = \mathbf{R}_y(q_3) \mathbf{R}_z(q_4) \quad (19)$$

Avec $L$ et $r$ sont les paramètres géométriques du manipulateur (longueur de la jambe et décalage de l'effecteur respectivement), et les autres variables sont les mêmes que dans l'équation (1). Comme dans le cas d'un manipulateur rigide, l'effecteur produit uniquement des mouvements de translation, les articulations passives sont soumis à des contraintes spécifiques $q_3 = -q_2$; $q_4 = -q_1$, qui sont implicitement utilisées dans le calcul du modèle géométrique direct et inverse. Toutefois, le modèle flexible permet des variations de toutes les articulations passives. En utilisant la modélisation de la raideur des membrures calculée avec les éléments finis et en appliquant notre méthode, nous avons calculé les matrices de raideur pour trois postures typiques du manipulateur et les résultats sont dans le tableau 1. On les compare ensuite avec un manipulateur avec des parallélogrammes.

## 4.2 Rigidité de l'Orthoglide avec des parallélogrammes

Avant d'évaluer la complaisance de manipulateur, nous devons évaluer la matrice de raideur des parallélogrammes. En utilisant les notations adoptées, nous pouvons écrire le modèle équivalent d'un parallélogramme

$$\mathbf{T}_{Plg} = \mathbf{R}_y(q_2) \cdot \mathbf{T}_x(L) \cdot \mathbf{R}_y(-q_2) \cdot \mathbf{V}_s(\theta_7, \dots \theta_{12}) \tag{20}$$

Où, par rapport au cas précédent, la troisième articulation passive est éliminé (il est implicitement admis que $q_3 = -q_2$). D'autre part, chaque parallélogramme peut être divisé en deux chaînes cinématiques sérielles (le «haut» et le «bas»)

$$\mathbf{T}_{haut} = \mathbf{T}_z(-d/2) \cdot \mathbf{R}_y(q+\Delta q_1^{up}) \cdot \mathbf{T}_x(L) \cdot \mathbf{V}_s(\theta_1^{up}, \dots \theta_6^{up}) \cdot \mathbf{R}_y(-q+\Delta q_2^{up}) \cdot \mathbf{T}_z(d/2) \tag{21}$$

$$\mathbf{T}_{bas} = \mathbf{T}_z(d/2) \cdot \mathbf{R}_y(q+\Delta q_1^{dn}) \cdot \mathbf{T}_x(L) \cdot \mathbf{V}_s(\theta_1^{dn}, \dots \theta_6^{dn}) \cdot \mathbf{R}_y(-q+\Delta q_2^{dn}) \cdot \mathbf{T}_z(-d/2) \tag{22}$$

Où $L$, $d$ sont les paramètres géométriques des parallélogrammes, $\Delta q_1^i, \Delta q_2^i, \ i \in \{up, dn\}$ sont les variations des articulations passives. Ainsi, les matrices de complaisance du parallélogramme peuvent être également obtenues à l'aide de notre technique qui donne une expression analytique,

$$\mathbf{K}_{Plg} = 2 \begin{bmatrix} K_{11} & 0 & 0 & 0 & 0 & 0 \\ 0 & K_{22} & 0 & 0 & 0 & K_{26} \\ 0 & 0 & 0 & 0 & 0 & 0 \\ \hline 0 & 0 & 0 & K_{44}+\dfrac{d^2 C_q^2 K_{22}}{4} & 0 & \dfrac{d^2 S_{2q} K_{22}}{8} \\ 0 & 0 & 0 & 0 & \dfrac{d^2 C_q^2 K_{11}}{4} & 0 \\ 0 & K_{26} & 0 & \dfrac{d^2 S_{2q} K_{22}}{8} & 0 & K_{66}+\dfrac{d^2 S_q^2 K_{22}}{4} \end{bmatrix} \quad \text{avec } C_q = \cos(q); S_q = \sin(q)$$

En utilisant ce modèle et en appliquant la technique proposée, on a calculé les matrices de rigidité pour trois postures particulières du manipulateur (voir tableau 1). Comme pour la comparaison avec l'utilisation des cardans, les parallélogrammes augment la raideur en torsion d'environ 10 fois. Cela justifie l'utilisation de cette architecture dans la conception du prototype de l'Orthoglide [15].

| TYPE D'ARCHITECTURE | POINT $Q_0$($x, y, z = 0.00$ mm) | | POINT $Q_1$($x, y, z = -73.65$ mm) | | POINT $Q_2$($x, y, z = +126.35$ mm) | |
|---|---|---|---|---|---|---|
| | $k_{tran}$ [MM/N] | $k_{rot}$ [RAD/N.MM] | $k_{tran}$ [MM/N] | $k_{rot}$ [RAD/N.MM] | $k_{tran}$ [MM/N] | $k_{rot}$ [RAD/N.MM] |
| MANIPULATEUR 3-PUU | $2.78 \cdot 10^{-4}$ | $20.9 \cdot 10^{-7}$ | $10.9 \cdot 10^{-4}$ | $24.1 \cdot 10^{-7}$ | $71.3 \cdot 10^{-4}$ | $25.8 \cdot 10^{-7}$ |
| MANIPULATEUR 3-PRPAR | $2.78 \cdot 10^{-4}$ | $1.94 \cdot 10^{-7}$ | $9.86 \cdot 10^{-4}$ | $2.06 \cdot 10^{-7}$ | $21.2 \cdot 10^{-4}$ | $2.65 \cdot 10^{-7}$ |

*Tableau 1 : Raideur de l'Orthoglide avec des jambes 3-PUU et 3-PRPaR*

## 5 Conclusion

Dans cet article, nous avons proposé une nouvelle méthode systématique pour le calcul de la matrice de rigidité des manipulateurs hyperstatiques parallèles. Cette méthode est basée sur une modélisation rigide avec des flexibilités localisées, dont les paramètres sont évalués par l'intermédiaire d'une modélisation par éléments finis et décrivent la complaisance en translation et en rotation ainsi que les couplages entre eux. Contrairement aux travaux précédents, notre méthode utilise une nouvelle stratégie pour écrire les équations cinématiques, qui considère en même temps les relations cinématiques et les conditions d'équilibre de chaque chaîne cinématique, puis regroupe les solutions partielles. Cela permet de calculer de la matrice de raideur des mécanismes hyperstatique dans n'importe quelle posture du manipulateur, y compris dans les postures singulières ou à leurs voisinages.

Un autre avantage de notre méthode est la simplicité des calculs qui exigent l'inversion de matrices de faible dimension par rapport à d'autres techniques. En outre, notre méthode ne nécessite pas l'élimination manuelle des flexibilités localisées redondantes associées aux articulations virtuelles, car cette opération est intrinsèquement incluse dans l'algorithme numérique. L'efficacité de notre méthode a été démontrée par des exemples d'application, qui traitent de l'analyse comparative de la raideur de deux manipulateurs parallèles de la famille de l'Orthoglide. Les résultats de nos calculs ont confirmé les avantages des architectures utilisant des parallélogrammes et ainsi validé la conception du prototype de l'Orthoglide. Une autre contribution importante de notre article est l'écriture d'un modèle analytique de la matrice de raideur des parallélogrammes, qui a été dérivée en utilisant la même méthodologie. Bien que nous ayons appliqué notre méthode à des mécanismes à 3 degrés de liberté de translation, notre méthode peut être étendue à d'autres architectures parallèles, composé de plusieurs chaînes cinématiques avec des articulations rotoïdes et prismatiques et des jambes différentes. Par la suite, nous allons appliquer notre méthode à des manipulateurs plus complexes comme la machine Verne [16], vérifier expérimentale notre modèle de rigidité sur le robot Orthoglide et prendre en compte l'influence de la gravité.

## Références